\documentclass[11pt]{article}
\usepackage{acl}
\usepackage{times}
\usepackage{latexsym}
\usepackage[T1]{fontenc}
\usepackage[utf8]{inputenc}
\usepackage{microtype}
\usepackage{booktabs}
\usepackage{graphicx}
\usepackage{amsmath}
\usepackage{amssymb}
\usepackage{algorithm}
\usepackage{algpseudocode}
\graphicspath{{figures/}}

\title{Audio-Native Speech Recognition with a Frozen \\ Discrete-Diffusion Language Model}

\author{
  Harsha Vardhan Khurdula$^{1}$ \quad Abhinav Kumar Singh$^{2}$ \quad Yoeven D Khemlani$^{2}$ \quad Vineet Agarwal$^{2}$ \\[2pt]
  Interfaze AI \\
  San Francisco, CA, USA \\
  \texttt{\{harsha, abhinav, yoeven, vineet\}@interfaze.ai}
}

\begin{document}
\maketitle

\begin{abstract}
Automatic speech recognition is dominated by autoregressive decoders that emit one token at a time. We ask whether a discrete diffusion language model can transcribe speech instead, refining a whole transcript in parallel over a small number of denoising steps. We train an audio-native interface for DiffusionGemma, a 26B mixture-of-experts model that generates text by uniform, random-token discrete diffusion rather than the absorbing-mask scheme common to recent diffusion language models. A frozen Whisper encoder supplies acoustic features, a lightweight projector maps them into the model embedding space, and low-rank adapters let the frozen backbone attend to the new modality. 

About 42M parameters are trained, which is 0.16 percent of the backbone. We find that the natural training objectives fail to ground the audio, because their gradient reaches the projector only through attention that has already dismissed it. A connectionist temporal classification loss applied through the frozen output head breaks this deadlock. The resulting model reaches 6.6 percent word error rate on LibriSpeech test-clean, transcribes in roughly eight parallel steps regardless of utterance length, and uses a single adapter trained on six languages, which we evaluate here on English, Hindi, and Mandarin.
\end{abstract}

\section{Introduction}

Speech recognition has been decoded the same way for a long time. An encoder reads the audio, and an attention decoder spells the transcript one token at a time, feeding each token back to itself to produce the next \citep{chan2016las,radford2023whisper}. The encoders have grown very strong, but the decoder is serial by construction, so a thirty second clip costs a few hundred sequential forward passes no matter how much hardware is available. Diffusion language models decode differently, starting from a canvas of noise and refining the whole sequence in parallel over a small number of denoising steps; they now reach quality competitive with autoregressive models of similar size \citep{nie2025llada}. The natural question is whether speech can be transcribed the same way, so that the cost of recognition is set by the number of denoising steps and not by the length of the transcript.

Two systems have shown that diffusion can transcribe. TransFusion denoised character sequences with multinomial diffusion \citep{baas2022transfusion}, and Whisfusion trained a masked-diffusion decoder on top of frozen Whisper features \citep{kwon2025whisfusion}. Both train a decoder specifically for the task, and both are English only. We ask a different question: can an existing, general-purpose diffusion language model be taught to hear, so that recognition runs through its own decoder with no new decoder trained at all?

We take DiffusionGemma, an open-weight 26B mixture-of-experts model that generates text by discrete diffusion (does not have audio support), and give it a native audio input. We keep the backbone frozen. We add a frozen Whisper encoder as an acoustic feature extractor \citep{radford2023whisper}, a small trainable projector that maps those features into the model embedding space, and low-rank adapters \citep{hu2022lora} so the frozen model can learn to attend to the new modality. The projector-into-language-model recipe follows the connectors that gave sight and hearing to text-only models \citep{liu2023llava,chu2023qwenaudio,rubenstein2023audiopalm}, applied here for the first time to a diffusion decoder. Only about 42M parameters train, which is 0.16 percent of the backbone. The result reaches 6.6 percent word error rate on LibriSpeech test-clean, converges in about eight parallel steps regardless of utterance length, and uses a single adapter trained on six languages.

The challenge lied in grounding, a frozen language model has never seen a spectrogram, so its embedding space has no notion of formants or phonemes, and the two obvious training objectives both route their signal through cross-attention. If attention has already decided the audio is noise, no gradient reaches the projector to make it useful, and the projector stays noise, which in turn gives attention no reason to change. We break this deadlock with a connectionist temporal classification loss applied directly through the frozen output head, which forces the audio embeddings to be linearly predictive of the transcript without asking attention to trust them first.

Our contributions are the following.
\begin{itemize}
  \item An audio-native pathway for a frozen discrete-diffusion language model, trained only through a projector and low-rank adapters, with no decoder trained from scratch.
  \item A diagnosis of the grounding failure and a direct-supervision fix that breaks the projector and attention deadlock in a few hundred steps.
  \item An empirical study of the accuracy and speed of diffusion decoding for speech, including a step count sweep that shows recognition cost decoupled from transcript length.
  \item A single adapter trained on six languages (English, German, French, Spanish, Hindi, Mandarin) and evaluated here on English, Hindi, and Mandarin, together with an analysis of where it trails autoregressive Whisper and why.
\end{itemize}

\section{Background}

\subsection{The diffusion backbone}
DiffusionGemma is a 26B mixture-of-experts model with 128 experts, top-8 routing, and roughly 4B active parameters per token. It has 30 transformer layers, a hidden size of 2816, and a vocabulary of 262144. It generates text by discrete diffusion rather than autoregression.

\begin{figure}[t]
  \centering
  \includegraphics[width=\columnwidth]{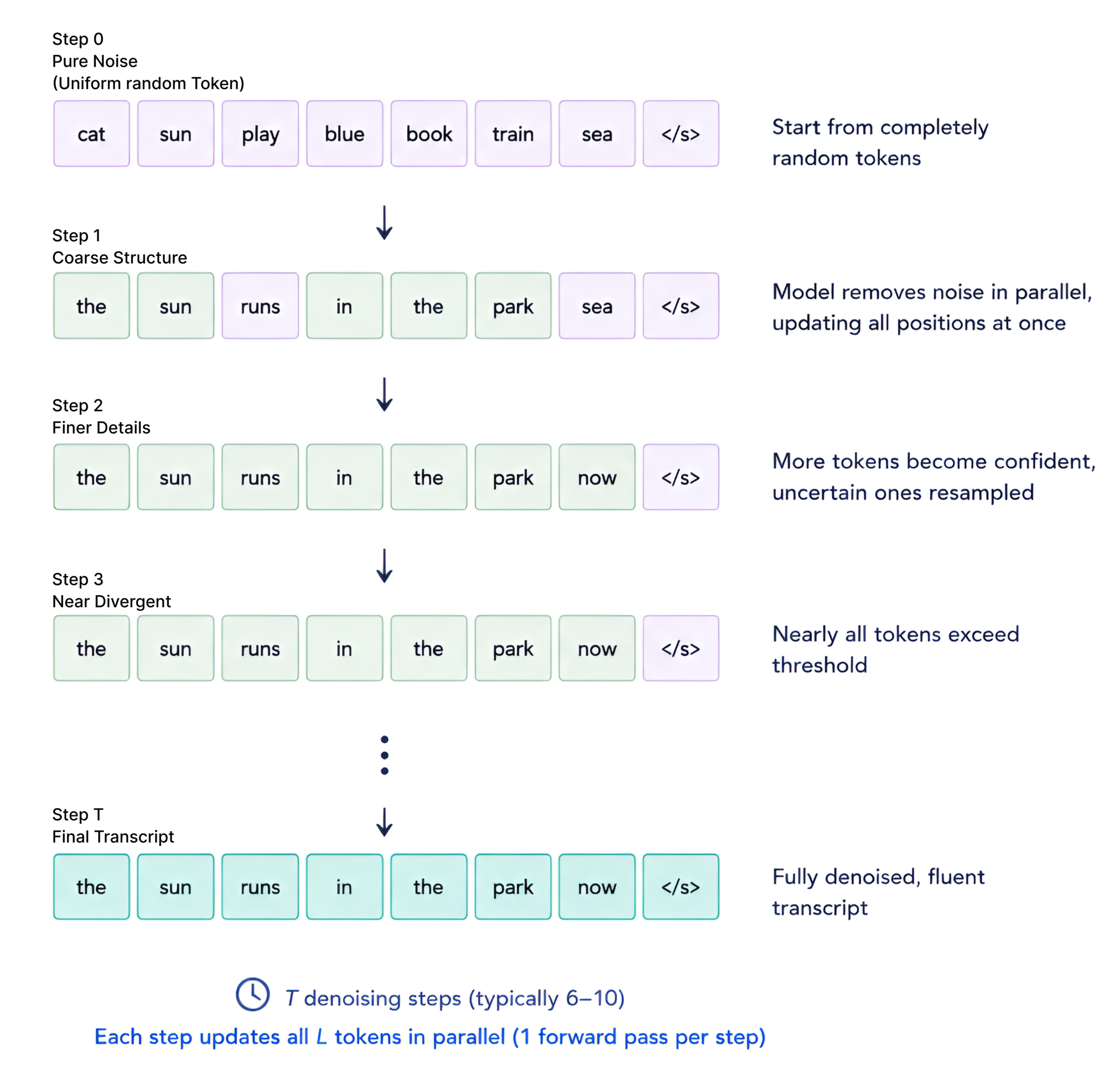}
  \caption{Uniform discrete diffusion in DiffusionGemma. Generation starts from a canvas of random vocabulary tokens and, over a few steps, keeps the confident predictions and renoises the rest until the canvas anneals into text. There is no absorbing mask state.}
  \label{fig:diffusion}
\end{figure}

Most recent diffusion language models use an absorbing state, where corruption replaces tokens with a special \texttt{<mask>} symbol and generation fills masks in \citep{austin2021d3pm,lou2024sedd,nie2025llada}. DiffusionGemma instead uses uniform, random-token diffusion. It starts from a fixed-length canvas, up to 256 slots, filled with random tokens drawn from the vocabulary. At each denoising step the model reads the whole canvas at once, keeps the low-entropy predictions it is confident about, and renoises the remaining positions back to fresh random tokens. After a few steps the noise anneals into text (Figure~\ref{fig:diffusion}). Training mirrors this process. A clean sequence is corrupted by replacing a fraction $\gamma$ of positions with uniform random tokens, and the model is trained to recover the originals at the corrupted positions. We formalize the process in Section~\ref{sec:diffusion}.


\subsection{Encoder-decoder structure}
Architecturally the model is an encoder-decoder with tied transformer weights. The encoder reads the prompt causally into a read-only key-value cache. The decoder refines the canvas with bidirectional self-attention while cross-attending to that cache. Multimodal inputs are injected by scattering projected features into placeholder token positions of the encoder input embeddings. The existing vision path uses an image token identifier for this. Out of the box the model accepts text, images, and video, and produces text. It does not accept audio and providing that modality is one of our key contributions in our work.

\subsection{Connecting an encoder to a language model}
A frozen language model can be given a new input modality by training a small connector that maps features from a modality encoder into the model embedding space, while the backbone stays fixed or is only lightly adapted. This recipe added vision to text-only models through a projector over a vision encoder \citep{liu2023llava}, and audio to language models through speech encoders and adapters \citep{chu2023qwenaudio,rubenstein2023audiopalm}, usually on top of open model families such as Gemma \citep{gemma2024}. Our audio pathway follows the same pattern. What is new is the decoder it feeds. Every connector we are aware of feeds an autoregressive model, whereas here the connector conditions a diffusion decoder that produces the transcript in parallel, which changes both how the model must be grounded and how it decodes.

\section{Method}

\begin{figure*}[t]
  \centering
  \includegraphics[width=\textwidth]{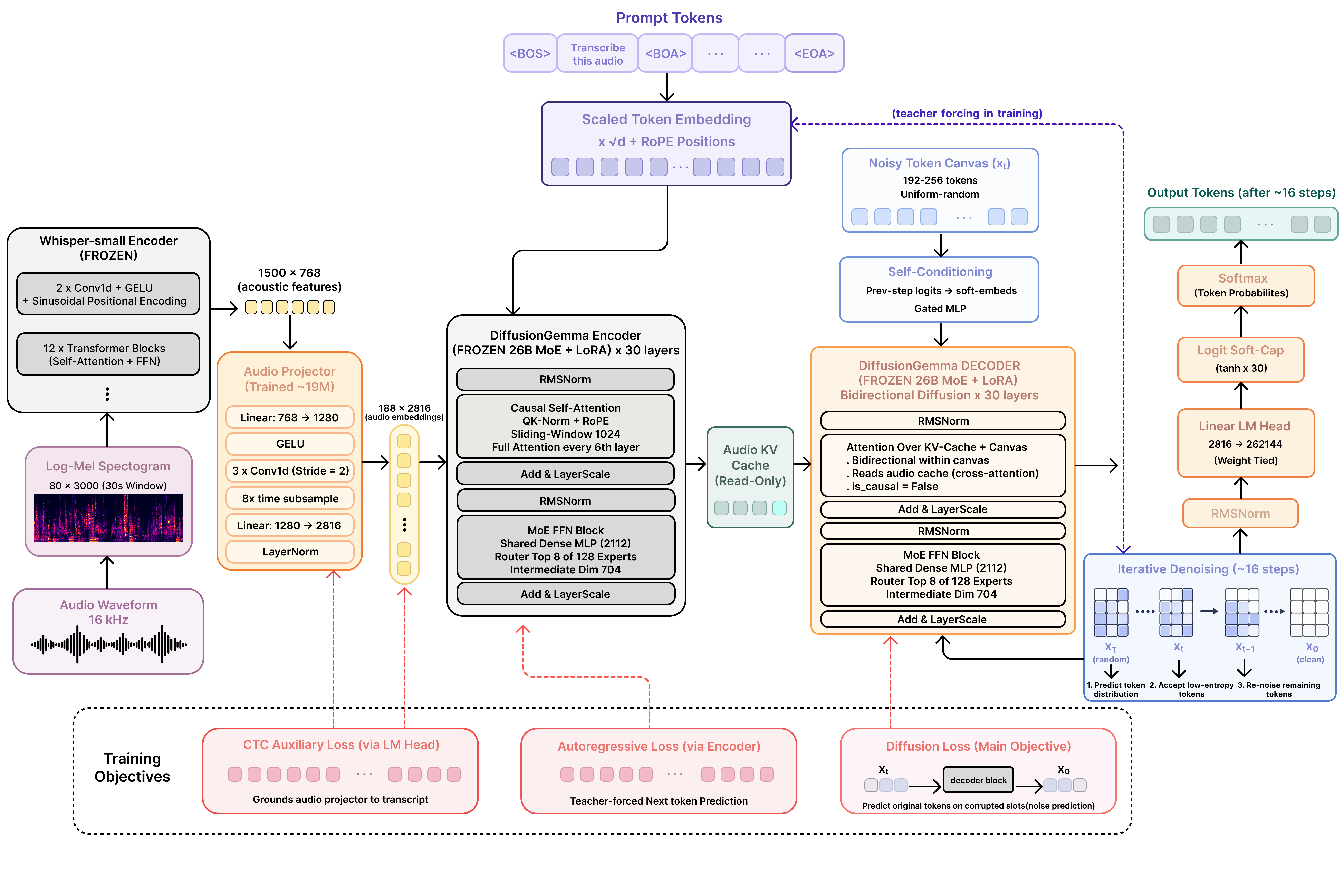}
  \caption{The audio-native pathway. A frozen Whisper encoder produces acoustic features, a trainable projector maps them to $T{=}188$ audio tokens in the model embedding space, and these are scattered into the placeholder positions of the encoder prompt. The frozen DiffusionGemma encoder reads the prompt into a key-value cache, and the decoder denoises the transcript canvas in parallel while attending to that cache. Three losses train the projector and the low-rank adapters. The diffusion loss and the autoregressive auxiliary loss route through attention, and the connectionist temporal classification loss is applied directly through the frozen output head.}
  \label{fig:arch}
\end{figure*}

\subsection{Notation and setup}
Let $\mathcal V$ be the vocabulary with $V=|\mathcal V|=262{,}144$. A transcript is a token sequence $\mathbf y=(y_1,\dots,y_M)$. We write it onto a fixed canvas of length $L$ (we use $L{=}256$ in training and $L{=}192$ at inference) as the clean target $\mathbf x_0=(\mathbf y,\textsc{eos},\textsc{pad},\dots)\in\mathcal V^{L}$. The audio is a log-mel spectrogram $\mathbf m\in\mathbb R^{80\times3000}$, since Whisper pads or trims to thirty seconds. We keep the DiffusionGemma backbone and the Whisper encoder frozen and train only a projector $\mathcal P_\phi$ and low-rank adapters $\theta$. We write $\sigma_\tau(z)=\tau\tanh(z/\tau)$ for the final logit softcap with $\tau{=}30$, and $W_U\in\mathbb R^{V\times d}$ for the frozen tied output head, with model width $d{=}2816$. The overall architecture is shown in Figure~\ref{fig:arch}.

\subsection{Audio pathway}
In our first attempt, we skipped the acoustic encoder entirely. Some unified Gemma models project raw waveforms straight into the embedding space, so we fed the model short audio frames and let the backbone infer the acoustics to test this approach. And it failed completely, as a frozen language model that has never seen audio cannot build an acoustic front end from gradient signal alone, and a shallow projector cannot do feature extraction on its own. Word error rate plateaued near 94 percent while the model produced fluent text unrelated to the audio.

The working design uses a frozen Whisper-small encoder \citep{radford2023whisper} as a feature extractor, not as a decoder. It carries the acoustic modeling that self-supervised representations \citep{baevski2020wav2vec2} and convolution-augmented transformers \citep{gulati2020conformer} established, which is the map from sound to linguistic features a frozen text model cannot learn on its own. The encoder produces $\mathbf a\in\mathbb R^{F\times d_w}$ with $F{=}1500$ frames and $d_w{=}768$. The projector maps these to audio tokens $\mathbf u\in\mathbb R^{T\times d}$,
\begin{align}
\mathbf h^{(0)} &= \mathrm{GELU}(\mathbf a\,W_{\text{in}}),\\
\mathbf h^{(l)} &= \mathrm{GELU}\!\big(\mathrm{Conv1d}^{(l)}(\mathbf h^{(l-1)})\big),\quad l=1,2,3,\\
\mathbf u &= \mathrm{LN}\!\big(\mathbf h^{(3)}\,W_{\text{out}}\big),
\end{align}
with $W_{\text{in}}\in\mathbb R^{d_w\times d_h}$, $W_{\text{out}}\in\mathbb R^{d_h\times d}$, hidden width $d_h{=}1280$, and convolutions of kernel 3, stride 2, and padding 1. Each convolution maps a length $n$ to $\lfloor(n-1)/2\rfloor+1$, so $1500\to750\to375\to188$ and $T{=}188$, about 6.25 tokens per second. The projector, together with the adapters introduced below, is the only trained part of the model.

\subsection{Conditioning the diffusion decoder on audio}
\label{sec:condition}
We build a fixed-length prompt
\[
\texttt{[BOS]}\ \text{instr}\ \texttt{[BOA]}\ \big(\texttt{<audio>}\big)^{\!\times T}\ \texttt{[EOA]},
\]
with audio placeholder positions $\mathcal A$, $|\mathcal A|{=}T$. We embed the prompt and overwrite the placeholders with the projected audio tokens,
\begin{equation}
\mathbf E_j=\begin{cases}\mathbf u_{\rho(j)} & j\in\mathcal A,\\[2pt] \mathrm{Emb}(c_j) & \text{otherwise,}\end{cases}
\end{equation}
where $\rho$ indexes projector rows, reusing the scatter mechanism the vision path uses. The frozen encoder, with adapters $\theta$, reads this into a key-value cache $\mathbf K=\mathrm{Enc}_\theta(\mathbf E)$, and the decoder cross-attends to $\mathbf K$ while denoising, as in Eq.~\ref{eq:denoiser}. The prompt length is constant because $T$ is fixed. For clips shorter than thirty seconds we zero the attention mask on the silence-pad audio positions so the encoder ignores them. The canvas holds the transcript, an end-of-sequence marker, and padding, and is the decoder target $\mathbf x_0$.

\subsection{Uniform discrete diffusion}
\label{sec:diffusion}
With the audio in the cache, we can state the generation process. Unlike the absorbing-state schemes used by most diffusion language models \citep{austin2021d3pm,lou2024sedd,nie2025llada}, DiffusionGemma corrupts toward uniform random tokens rather than a mask. A single scalar $\gamma\in[0,1]$ sets the corruption level and plays the role a timestep plays in continuous diffusion. Corruption acts independently at each position,
\begin{equation}
q(x_\gamma^i = k \mid x_0^i;\gamma) = (1-\gamma)\,\mathbf{1}[k=x_0^i] + \frac{\gamma}{V},
\label{eq:forward}
\end{equation}
which is the cumulative uniform D3PM kernel $\bar Q_\gamma=(1-\gamma)\,I+\tfrac{\gamma}{V}\mathbf 1\mathbf 1^{\!\top}$ \citep{austin2021d3pm}. Equivalently, draw $b_i\sim\mathrm{Bernoulli}(\gamma)$ and $r_i\sim\mathrm{Unif}(\mathcal V)$ and set $x_\gamma^i=(1-b_i)\,x_0^i+b_i\,r_i$. There is no special mask symbol, so a corrupted position holds a real, random vocabulary token.

The model is trained in the $\mathbf x_0$ parameterization. Given the corrupted canvas and the audio conditioning $\mathbf c$ carried by the cache $\mathbf K$, it predicts the clean token at each position,
\begin{equation}
\begin{aligned}
p_{\theta,\phi}(x_0^i\mid \mathbf x_\gamma,\mathbf c)&=\mathrm{softmax}\!\big(\sigma_\tau(W_U\,\mathbf h_i)\big),\\
\mathbf h&=\mathrm{Dec}_\theta(\mathbf x_\gamma,\mathbf K).
\end{aligned}
\label{eq:denoiser}
\end{equation}

\subsection{Training objectives}
Three losses share one projector and one set of adapters.

\paragraph{Diffusion loss.} This is the model's own objective, the reweighted $\mathbf x_0$-prediction surrogate of the discrete-diffusion variational bound \citep{austin2021d3pm,ho2020ddpm}. Let $\mathcal C$ be the corrupted, non-pad positions. Taking the expectation over $\gamma\sim\mathrm{Unif}(0,1)$, the data, and the corruption,
\begin{equation}
\mathcal L_{\text{diff}}=\mathbb E\!\left[\frac{w(\gamma)}{|\mathcal C|}\sum_{i\in\mathcal C}-\log p_{\theta,\phi}(x_0^i\mid \mathbf x_\gamma,\mathbf c)\right],
\label{eq:diff}
\end{equation}
with an optional importance weight $w(\gamma)\in\{1,\,1/\gamma\}$. The loss is applied only at corrupted positions. Because the objective is bidirectional and the frozen text prior explains much of the visible clean context, the gradient asking the model to use audio is diluted (Section~\ref{sec:grounding}). We therefore force $\gamma{=}1$ on a fraction of batches, where the audio is the only source of information.

\paragraph{Autoregressive auxiliary loss.} The encoder is causal, so we can teacher-force the transcript through it and read a standard next-token loss,
\begin{equation}
\mathcal L_{\text{ar}}=\mathbb E\Big[\frac1M\sum_{i=1}^{M}-\log\,\mathrm{softmax}\!\big(\sigma_\tau(W_U\mathbf g_i)\big)_{y_i}\Big],
\label{eq:ar}
\end{equation}
where $\mathbf g_i$ is the encoder state that predicts $y_i$ from $\mathbf y_{<i}$ and the audio. Because the encoder shares weights with the diffusion decoder and reads the same audio tokens, this dense per-token gradient transfers to parallel denoising.

\paragraph{CTC loss.} We decode the projector rows directly through the frozen output head, with no softcap, and align them to the transcript with connectionist temporal classification \citep{graves2006ctc}. With per-frame distributions $\mathbf q_t=\mathrm{softmax}(W_U\mathbf u_t)$ and the blank symbol at index 0,
\begin{equation}
\mathcal L_{\text{ctc}}=-\log\!\!\sum_{\pi\in\mathcal B^{-1}(\mathbf y)}\;\prod_{t=1}^{T}\mathbf q_t(\pi_t),
\label{eq:ctc}
\end{equation}
where $\mathcal B$ collapses repeats and removes blanks. Feasibility needs $T\ge M$, so we cap targets to the number of real audio tokens. The full objective is
\begin{equation}
\mathcal L=\mathcal L_{\text{diff}}+\lambda_{\text{ar}}\,\mathcal L_{\text{ar}}+\lambda_{\text{ctc}}\,\mathcal L_{\text{ctc}}.
\label{eq:total}
\end{equation}

\paragraph{Low-rank adaptation.} We add LoRA adapters \citep{hu2022lora} of rank 16 and scaling 32 on the query, key, value, and output projections of both the encoder and the decoder self-attention. The encoder adapters let the model fold audio into the cache, and the decoder adapters let it attend to that cache during denoising. The experts and the vision tower stay frozen. The projector and the adapters together are 42.3M trainable parameters, which is 0.16 percent of the backbone.

\subsection{Why grounding fails, and how CTC fixes it}
\label{sec:grounding}
With only the diffusion and autoregressive losses, training stalled in a way that took us a while to understand. The curves fell for a few hundred steps and then went flat. The diffusion loss settled near 8 and the autoregressive loss sat around 4.5, close to chance, and neither moved for thousands of steps.

Both losses reach the projector only through attention over the audio positions. Let $\alpha_t$ be the aggregate attention mass the model places on audio token $t$. To first order the gradient the attention-routed losses send to $\mathbf u_t$ scales with that mass,
\begin{equation}
\frac{\partial \mathcal L_{\text{diff}}}{\partial \mathbf u_t}\;\sim\;\alpha_t\,W_V^{\!\top}\boldsymbol\delta_t,
\label{eq:gradstall}
\end{equation}
where $\boldsymbol\delta_t$ is the backpropagated error at the attended positions and $W_V$ the value projection. At initialization $\mathbf u_t$ is random, the frozen prior explains the visible clean context, and the model learns $\alpha_t\to0$, which sends this gradient to zero. The gradient that would raise $\alpha_t$ is also small, because $\mathbf u_t$ carries nothing worth attending to. Both directions vanish at the same fixed point. The projector and the attention each wait for the other to move first.

CTC removes the $\alpha_t$ factor. Its gradient to the projector is the linear-head cross-entropy gradient composed with the CTC forward-backward occupancy $\boldsymbol\rho_t$, the posterior probability of each label at frame $t$,
\begin{equation}
\frac{\partial \mathcal L_{\text{ctc}}}{\partial \mathbf u_t}=W_U^{\!\top}\big(\mathbf q_t-\boldsymbol\rho_t\big),
\label{eq:gradctc}
\end{equation}
which is nonzero whatever the attention does. It pushes $\mathbf u_t$ into the region of embedding space where $W_U\mathbf u_t$ predicts transcript tokens, with no attention required. Once $\mathbf u_t$ carries content, $\alpha_t$ grows and the gradients in Eq.~\ref{eq:gradstall} revive. Adding the term broke the plateau on our first attempt: the CTC loss fell from 24 to 8.6 in about 300 steps and held-out token accuracy climbed off the floor. The CTC head is a training-time scaffold, which we drop at inference so that recognition runs entirely through the diffusion decoder.

\subsection{Inference as parallel denoising}
Whisper's encoder is fixed to a thirty second window, so longer audio is split at silence. We take the quietest short frame near a thirteen second target inside a three second search window and cut there, so no word is split and the non-overlapping segments concatenate with no stitching. Each segment is denoised on a canvas of $L{=}192$ slots by Algorithm~\ref{alg:sample}.

The sampler is the base model's uniform-diffusion process, conditioned on audio. Each step recomputes the denoiser logits over the whole canvas, accepts the positions whose predictive entropy $H(\mathbf p_i)$ falls below a bound $\eta$, and renoises the rest to fresh uniform tokens. A linear temperature schedule anneals from 0.8 down to 0.4, self-conditioning \citep{chen2023selfcond} feeds the previous prediction forward, and a stability-and-confidence criterion stops early once the canvas settles. Optional classifier-free guidance \citep{ho2022cfg} combines a conditional pass with an unconditional pass that hides the audio cache, $\boldsymbol\ell=\boldsymbol\ell^{u}+w(\boldsymbol\ell^{c}-\boldsymbol\ell^{u})$, to sharpen the effect of the audio without retraining. A short post-processing pass removes repeated words and n-grams, the canvas-fill artifacts a diffusion decoder produces when the transcript is shorter than the canvas.

\begin{algorithm}[t]
\caption{Audio-conditioned parallel denoising}
\label{alg:sample}
\begin{algorithmic}[1]
\State \textbf{input:} audio $\mathbf a$, prompt $\mathbf c$, steps $S$, guidance $w$, bound $\eta$
\State $\mathbf K \gets \mathrm{Enc}_\theta(\mathrm{scatter}(\mathrm{Emb}(\mathbf c),\,\mathcal P_\phi(\mathbf a)))$
\State $\mathbf x \gets$ uniform random in $\mathcal V^{L}$
\For{$s = 1$ \textbf{to} $S$}
  \State $\boldsymbol\ell \gets \sigma_\tau\!\big(W_U\,\mathrm{Dec}_\theta(\mathbf x,\mathbf K)\big)$
  \If{$w \ne 1$}
     \State $\boldsymbol\ell^{u} \gets \sigma_\tau\!\big(W_U\,\mathrm{Dec}_\theta(\mathbf x,\mathbf K^{\setminus \text{audio}})\big)$
     \State $\boldsymbol\ell \gets \boldsymbol\ell^{u} + w\,(\boldsymbol\ell - \boldsymbol\ell^{u})$
  \EndIf
  \State $\mathbf p \gets \mathrm{softmax}(\boldsymbol\ell/\text{temp}(s))$;\; $\hat{\mathbf x}\sim\mathrm{Multinomial}(\mathbf p)$
  \State keep $\hat x_i$ where $H(\mathbf p_i)<\eta$; renoise the rest to fresh uniform tokens
  \State \textbf{if} $\arg\max\boldsymbol\ell$ stable and confident \textbf{then break}
\EndFor
\State \textbf{return} $\arg\max\boldsymbol\ell$, decoded up to \textsc{eos}
\end{algorithmic}
\end{algorithm}

\paragraph{Cost.} Autoregressive decoding needs $M$ sequential decoder passes to emit $M$ tokens \citep{vaswani2017attention}. Diffusion decoding needs at most $S$ passes over the whole canvas in parallel, plus one encoder pass, and doubles the decoder passes only when guidance is on. The cost is set by $S$ and is independent of the transcript length $M$. With $S\approx8\ll M$, and the flat accuracy curve of Table~\ref{tab:steps}, this is the source of the speedup.

\section{Experimental Setup}

\paragraph{Data.} Stage one, we train the English model on the 100 hour clean subset of LibriSpeech \citep{panayotov2015librispeech}. The multilingual model warm-starts from the English checkpoint and continues on FLEURS \citep{conneau2023fleurs} across six languages, English, German, French, Spanish, Hindi, and Mandarin. A further stage adds VoxPopuli parliamentary speech \citep{wang2021voxpopuli}. The model has seen roughly 219 hours of audio in total.

\paragraph{Optimization.} We use AdamW with a learning rate of $1\mathrm{e}{-3}$, betas of 0.9 and 0.95, and weight decay of 0.01, with 100 warmup steps and a cosine decay to one tenth of the peak. We clip gradients at norm 1.0. The backbone runs in bfloat16 and the projector in float32 for stable updates. Batches hold 4 utterances with 2 gradient accumulation steps, reduced to 2 and 6 on the longer VoxPopuli material. All runs use a single H100.

\paragraph{Evaluation.} We report word error rate with the Whisper text normalizer applied to both reference and hypothesis, which is the Open-ASR and Artificial Analysis convention. For Hindi and Mandarin we report character error rate, because word error rate penalizes scripts whose word boundaries do not match the reference segmentation and overstates the error. We also report throughput as a multiple of real time.

\paragraph{Configuration.} Table~\ref{tab:hparams} lists the settings used for the reported runs.

\begin{table}[t]
\centering
\small\setlength{\tabcolsep}{4pt}
\begin{tabular}{@{}lp{4.9cm}@{}}
\toprule
Setting & Value \\
\midrule
Optimizer        & AdamW $(0.9,0.95)$, weight decay $0.01$ \\
Learning rate    & $1\mathrm{e}{-3}$, 100 warmup, cosine to $0.1\times$ \\
Gradient clip    & $1.0$ \\
Batch $\times$ accum & $4\times2$ (Libri/FLEURS), $2\times6$ (VoxPopuli) \\
Precision        & bf16 backbone, fp32 projector \\
Canvas $L$       & 256 train, 192 inference \\
Subsample / tokens & $8\times$ / $T{=}188$ \\
LoRA             & $r{=}16$, $\alpha{=}32$, dropout $0.05$, enc+dec attn \\
Loss weights     & $\lambda_{\text{ar}}{=}1$, $\lambda_{\text{ctc}}{=}1$, $w(\gamma){=}1$ \\
Inference        & 8 steps, $\eta{=}0.1$, guidance $w{=}1$, temp $0.8\!\to\!0.4$ \\
\bottomrule
\end{tabular}
\caption{Training and inference configuration for the reported runs.}
\label{tab:hparams}
\end{table}

\section{Results}

\subsection{English recognition and grounding dynamics}
On LibriSpeech test-clean, English word error rate walked down over ten epochs, from roughly 90 percent early in training to about 50 percent, then to 14.6 percent, and finally to 6.6 percent at the last checkpoint ($n{=}100$ test-clean utterances). The two intermediate points come from small held-out samples and are approximate; the final number is on the full evaluation set. This is single-digit word error rate from a frozen 26B backbone, a frozen Whisper encoder, and about 42M trainable parameters.

Grounding was not monotone in the metrics. Token accuracy reached 0.50 and CTC loss kept falling while the model still said nothing useful. A manual decode at 0.4 epochs showed the greedy CTC output emitting the unigram prior, the most frequent tokens in frequency order such as ``the of to he'', with no relation to the audio, even though the loss curve looked healthy. Only repeated exposure over many passes brought the projector and the attention into a configuration that actually transcribes, so we ran a manual decode every half epoch. As training progressed the outputs moved from repeated fragments to fluent text with occasional acoustic slips, such as ``tents'' for ``tense'', the error profile of a model transcribing from the audio rather than falling back on its text prior.

Table~\ref{tab:grounding} shows the same run in numbers. The autoregressive loss stays near chance and token accuracy hovers around 0.4 to 0.5 through the first epoch, then both move sharply once the projector grounds. Without the CTC term the same configuration did not ground within the compute we ran, which is the failure Section~\ref{sec:grounding} analyzes; we observed this during development rather than as a controlled ablation.

\begin{table}[t]
\centering
\begin{tabular}{lrr}
\toprule
Step & Val AR loss & Val token acc \\
\midrule
25            & 5.01 & 0.42 \\
300           & 4.62 & 0.44 \\
900           & 4.45 & 0.50 \\
2808 (ep.\ 2) & 0.34 & 0.85 \\
\bottomrule
\end{tabular}
\caption{Grounding dynamics on a held-out set of about 32 utterances. The autoregressive loss stays near chance and token accuracy plateaus while the model is ungrounded, then both improve sharply. Over the same span the CTC loss falls from 24 to 8.6 and then below 0.1.}
\label{tab:grounding}
\end{table}

\subsection{Multilingual recognition}
Warm-starting from English and training on FLEURS gave one adapter covering all six languages. We report the three for which we ran held-out evaluation, English, Hindi, and Mandarin. German, French, and Spanish are in the training mix but were not separately evaluated here. Adding VoxPopuli traded a small amount of read-speech accuracy for a gain on conversational and accented audio, a movement along the accuracy frontier rather than a free improvement. Table~\ref{tab:multiling} reports the adapter under the Whisper normalizer.

\begin{table}[t]
\centering
\begin{tabular}{llrr}
\toprule
Benchmark & Metric & Score & $n$ \\
\midrule
FLEURS English   & WER & 15.7\% & 150 \\
VoxPopuli English & WER & 18.5\% & 1000 \\
FLEURS Hindi     & CER & 15.8\% & 300 \\
FLEURS Mandarin  & CER & 29.6\% & 300 \\
\bottomrule
\end{tabular}
\caption{Multilingual adapter, Whisper-normalized, with evaluation set sizes $n$. Mandarin word error rate reads 40 percent under word segmentation, while the character error rate on the same transcripts is 29.6 percent.}
\label{tab:multiling}
\end{table}

\subsection{Parallel decoding}
Because the decoder refines the entire canvas at once, transcription cost is set by the number of denoising steps and not by the length of the transcript. Table~\ref{tab:steps} sweeps the step count on FLEURS English. The curve is close to flat. Moving from 8 steps to 48 buys about half a point of word error rate and costs roughly three times the latency. The model converges in about eight parallel passes. For a ten second clip this is on the order of one second of model time, and it does not grow with how much the speaker says.

\begin{table}[t]
\centering
\begin{tabular}{lrr}
\toprule
Steps & FLEURS-en WER & Speed \\
\midrule
8  & 15.7\% & 14.9$\times$ \\
16 & 15.6\% & 10.3$\times$ \\
32 & 15.2\% & 6.5$\times$ \\
48 & 15.6\% & 4.7$\times$ \\
\bottomrule
\end{tabular}
\caption{Word error rate and throughput as a multiple of real time against the number of denoising steps (FLEURS English, $n{=}150$).}
\label{tab:steps}
\end{table}

\subsection{Comparison with diffusion and autoregressive systems}
Table~\ref{tab:diffasr} places our LibriSpeech test-clean result beside other diffusion and non-autoregressive systems. These are the numbers reported in the respective papers, not controlled reruns, and the systems differ in encoder, training data, tokenization, and decoding, so the table is context rather than a head-to-head comparison. We place our result next to TransFusion \citep{baas2022transfusion} and Whisfusion \citep{kwon2025whisfusion} for reference, not to claim a controlled win.

\begin{table}[t]
\centering
\begin{tabular}{llr}
\toprule
Model & Approach & WER \\
\midrule
TransFusion & multinomial diffusion & $\sim$6--7\% \\
Whisfusion  & masked diffusion & 8.3\% \\
Ours        & frozen diffusion LM & \textbf{6.6\%} \\
\bottomrule
\end{tabular}
\caption{Diffusion and non-autoregressive speech recognition on LibriSpeech test-clean. Baseline numbers are as reported by their authors under differing encoders, data, and decoding, and are not controlled reruns; only our row ($n{=}100$) is measured in this work. TransFusion is a character-level proof of concept.}
\label{tab:diffasr}
\end{table}

Against autoregressive Whisper, which remains the strongest baseline, our model trails, as Table~\ref{tab:whisper} shows. All numbers use the Whisper normalizer. The gap is largest on FLEURS and VoxPopuli and smallest on LibriSpeech, the one domain where we had a hundred clean hours.

\begin{table}[t]
\centering
\small\setlength{\tabcolsep}{4pt}
\begin{tabular}{lrrr}
\toprule
Benchmark & Ours & W-small & W-large-v3 \\
\midrule
LibriSpeech clean & 6.6\%  & $\sim$3.4\% & $\sim$2.0\% \\
FLEURS-en         & 15.7\% & $\sim$9--10\% & $\sim$4--5\% \\
VoxPopuli-en      & 18.5\% & $\sim$9--11\% & $\sim$7--10\% \\
\bottomrule
\end{tabular}
\caption{Context against autoregressive Whisper. The Whisper columns are published ranges, not controlled reruns; only the Ours column is measured here (LibriSpeech $n{=}100$, FLEURS-en $n{=}150$, VoxPopuli-en $n{=}1000$).}
\label{tab:whisper}
\end{table}

\subsection{What is the bottleneck}
We ran ablations to locate the source of the residual error, because the answer decides whether the method is broken or simply undertrained. We ruled out one suspect at a time.

First, encoder quality. We swapped the Whisper-small front end for Whisper-large-v3 and retrained on the same 100 hour set. Word error rate at matched data moved from 14.75 percent at three epochs to 13.15 at six and 8.95 at eight, which is roughly even with the small-encoder run at the same data scale. A stronger encoder did not move the floor at this data budget, so the encoder is not the current limit.

Second, audio resolution. We reduced the subsample factor below eight to give the projector more audio tokens per second. This regressed to 44.7 percent word error rate rather than improving, so token resolution is not the limit either, and the eight-fold subsample is not starving the decoder.

Third, decode steps. The step sweep in Table~\ref{tab:steps} is flat, so the sampler is not leaving accuracy on the table.

With encoder quality, resolution, and step count ruled out, the most likely remaining explanation is data scale and adapter alignment. Whisper-large-v3 was trained on a few million hours and Whisper-small on about 680 thousand. This model has seen roughly 219 hours, three orders of magnitude fewer, and it already reached 6.6 percent on the single domain where it had a hundred clean hours. The frontend ablation points the same way, since a stronger encoder did not help at this data budget, which is what one expects when the bottleneck is data rather than features. The evidence is consistent with a data bottleneck rather than a broken method, but we cannot establish this without a scaling curve, which we leave to future work.

\section{Related Work}

\paragraph{How speech recognition decodes.} End-to-end recognition grew from two ideas. Connectionist temporal classification aligns a frame sequence to a shorter label sequence without explicit segmentation \citep{graves2006ctc}. Attention-based sequence-to-sequence models instead let a decoder attend over the encoded audio and spell the transcript \citep{chan2016las}, a pattern the transformer generalized \citep{vaswani2017attention}. Most progress since then has come from stronger encoders, including self-supervised representations \citep{baevski2020wav2vec2} and convolution-augmented transformers \citep{gulati2020conformer}, and from scale and weak supervision \citep{radford2023whisper}. In all of these the transcript is produced one token at a time.

\paragraph{Non-autoregressive and diffusion text generation.} Parallel generation began with non-autoregressive translation, which emits the whole target at once \citep{gu2018nat}. Diffusion models, developed for continuous data \citep{sohldickstein2015diffusion,ho2020ddpm}, were brought to language both by embedding tokens in a continuous space \citep{li2022diffusionlm,gong2023diffuseq} and by defining the corruption directly on discrete tokens \citep{austin2021d3pm,lou2024sedd}. Self-conditioning improved discrete diffusion by feeding the previous prediction back into the next step \citep{chen2023selfcond}, a technique our sampler uses. Large masked-diffusion language models now reach quality competitive with autoregressive models of similar size \citep{nie2025llada}. DiffusionGemma sits in this family but corrupts toward uniform random tokens rather than an absorbing mask.

\paragraph{Diffusion speech recognition and audio language models.} Two systems put diffusion decoding directly on speech. TransFusion transcribed characters with multinomial diffusion as a proof of concept \citep{baas2022transfusion}, and Whisfusion trained a masked-diffusion decoder over frozen Whisper features \citep{kwon2025whisfusion}. Both build that decoder from scratch, whereas we reuse one that already exists. Separately, a large body of work gives new modalities to frozen or lightly adapted language models through a trained connector, for vision \citep{liu2023llava} and for audio \citep{chu2023qwenaudio,rubenstein2023audiopalm}, often on top of open model families such as Gemma \citep{gemma2024}. Our work joins these two threads. We reuse an off-the-shelf diffusion language model with no new decoder, connect a frozen speech encoder through a projector and low-rank adapters \citep{hu2022lora}, ground it with connectionist temporal classification \citep{graves2006ctc}, decode with uniform random-token diffusion rather than absorbing masks, and train one adapter on six languages.

\section{Limitations}
The model trails autoregressive Whisper on every benchmark we tested, by the largest margin on multilingual read speech. The evidence points to data scale as the cause rather than architecture, but that remains a hypothesis until we train on substantially more audio. The Whisper encoder fixes the input window at thirty seconds, so real streaming is not possible and long audio is handled by silence segmentation, which can err when speech has no clear pauses. The diffusion decoder can produce repetition artifacts on short utterances, which we currently remove with a post-processing pass rather than at the source. Finally, our comparisons to Whisper use published ranges under the same normalizer rather than a single controlled re-run of every baseline, so the Whisper columns should be read as reference points rather than exact head-to-head numbers.

\section{Conclusion}
A frozen language model can learn to hear if the audio projector is supervised directly rather than only through attention. Loss curves and token accuracy will look healthy well before the model is actually grounded, so manual decoding matters during training. Once grounded, diffusion decoding makes the length of what a speaker said stop mattering for cost, because recognition converges in about eight parallel steps. The grounding recipe, the CTC unlock, and the diffusion decoder all work today; what the model still lacks is exposure to more hours of audio.


\bibliographystyle{acl_natbib}

\end{document}